\documentclass{article}

\usepackage{PRIMEarxiv}

\usepackage[colorlinks, citecolor=blue,breaklinks = true]{hyperref}       
\usepackage{url}            
\usepackage{booktabs}       
\usepackage{amsfonts}       
\usepackage{nicefrac}       
\usepackage{microtype}      
\usepackage{lipsum}
\usepackage{fancyhdr}       
\usepackage{epsfig}
\usepackage{graphicx}       
\graphicspath{{media/}}     
\usepackage{amsmath}
\usepackage{amssymb} 
\usepackage{multirow}
\usepackage{diagbox}
\usepackage{appendix}
\usepackage{enumerate}

\usepackage[linesnumbered,algo2e]{algorithm2e}
\usepackage{algorithm,algcompatible}

\newcommand{\ie}{\textit{i}.\textit{e}.}
\newcommand{\eg}{\textit{e}.\textit{g}.}
\newlength\savewidth\newcommand\shline{\noalign{\global\savewidth\arrayrulewidth
  \global\arrayrulewidth 1.5pt}\hline\noalign{\global\arrayrulewidth\savewidth}}
  
\newlength\lighterwidth
\newcommand{\tabincell}[2]{\begin{tabular}{@{}#1@{}}#2\end{tabular}}

\usepackage{caption}
\usepackage{subcaption}

\title{Forget Less, Count Better: A Domain-Incremental Self-Distillation Learning Benchmark for Lifelong Crowd Counting
}

\author{
  Jiaqi Gao$^1$, Jingqi Li$^1$, Hongming Shan$^1$, Yanyun Qu$^2$, James Z. Wang$^3$, Fei-Yue Wang$^4$, Junping Zhang$^1$ \\ \\
  $^1$Fudan University, Shanghai, China \\ 
  $^2$Xiamen University, Xiamen, China \\
  $^3$The Pennsylvania State University, Pennsylvania, USA\\
  $^4$Chinese Academy of Sciences, Beijing, China\\
  \texttt{\{jqgao20, lijq20, hmshan, jpzhang\}@fudan.edu.cn,} 
  \\\texttt{yyqu@xmu.edu.cn, jwang@ist.psu.edu, feiyue.wang@ia.ac.cn}
}

\begin{document}
\maketitle

\begin{abstract}
Crowd counting has important applications in public safety and pandemic control. A robust and practical crowd counting system has to be capable of continuously learning with the new incoming domain data in real-world scenarios instead of fitting one domain only. Off-the-shelf methods have some drawbacks when handling multiple domains: (1) the models will achieve limited performance (even drop dramatically) among old domains after training images from new domains due to the discrepancies of intrinsic data distributions from various domains, which is called catastrophic forgetting; (2) the well-trained model in a specific domain achieves imperfect performance among other unseen domains because of the domain shift; and (3) it leads to linearly increasing storage overhead, either mixing all the data for training or simply training dozens of separate models for different domains when new ones are available. To overcome these issues, we investigated a new crowd counting task in the incremental domains training setting called Lifelong Crowd Counting. Its goal is to alleviate the catastrophic forgetting and improve the generalization ability using a single model updated by the incremental domains. Specifically, we propose a self-distillation learning framework as a benchmark~(\textbf{F}orget \textbf{L}ess, \textbf{C}ount \textbf{B}etter, or \textbf{FLCB}) for lifelong crowd counting, which helps the model sustainably leverage previous meaningful knowledge for better crowd counting to mitigate the forgetting when the new data arrive. In addition, a new quantitative metric, normalized backward transfer~(nBwT), is developed to evaluate the forgetting degree of the model in the lifelong learning process. Extensive experimental results demonstrate the superiority of our proposed benchmark in achieving a low catastrophic forgetting degree and strong generalization ability.
\end{abstract}

\begin{figure*}
  \includegraphics[width=\textwidth]{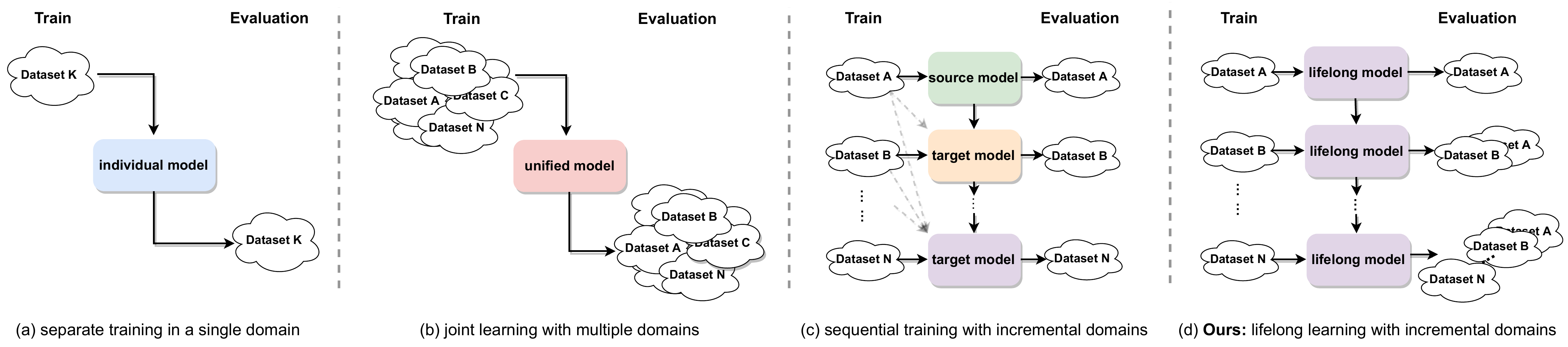}
\caption{The conceptual differences of four training paradigms: (a) Directly training an individual model for each dataset. (b) Training a unified model by mixing all datasets from different domains. (c) Leveraging previous data or models to improve the performance on the target domain dataset. (d) Ours: lifelong learning with incremental domains to improve the performance among all domains. In (c), the dotted lines indicate that the past domain data may be used repeatedly to improve the performance in the target domain dataset. In (d), our proposed FLCB model does not replay any previous domain data and evaluates all domain datasets at the training stage. Without storing previous domain data, our proposed lifelong model~(FLCB) itself can still sustainably handle the crowd counting problem among multiple domains, being updated by the new available domain dataset only.}
  \label{figures: introduction}
\end{figure*}

\section{Introduction}

Crowd counting is to predict the number of people in an image or a video sequence. Accurate crowd counting for crowded scenes has important applications such as traffic control, preventing stampedes from occurring, and estimating participation in large public events like parades. For example, during a pandemic, authorities may need to maintain social distancing for public spaces to minimize the risk of infection. Thus, crowd counting systems are usually deployed in multiple diverse scenarios, such as malls, museums, squares, and public squares. For one site, the running system is expected to continually handle the non-stationary data with different densities, illumination, occlusion, and various head scales. For multiple sites, the system should also consider dozens of scenes and perspective information.

As data are increasingly produced and labeling is time-consuming, the new domain data available for training are usually collected and labeled incrementally. We may ask: \emph{how to sustainably handle the crowd counting problem among multiple domains using a single model when the newly available domain data arrive?} We try to find the most potential solution to this question from the following aspects.

Currently, most crowd counting approaches~\cite{zhang2016single, babu2017switching,sindagi2017generating,jiang2019learning,cao2018scale, chen2019scale,song2021choose,liu2019context,li2018csrnet,tian2019padnet,bai2020adaptive,ma2019bayesian,tan2019crowd,ma2020learning} concentrate on training an independent model for each single domain dataset. They heavily rely on the \emph{i.i.d.} assumption that images from both the training set and test set are independently and identically distributed. Although producing promising counting performance in the corresponding domain, such a training strategy, shown in Figure~\ref{figures: introduction}(a), has drawbacks in dealing with multiple and incremental new datasets, which are common in the real world; \eg, when limited labeled data from a new site are available before applying the model at the site. One drawback is that these separately trained models often have low generalization ability when dealing with new, unseen domain data due to the domain shift evidenced in Table~\ref{results: singledomain}. Another is that saving multiple different sets of trained parameters from distinct domains for inference is not economical when deploying them to hundreds of thousands of real-world sites. Training a shared and universal model from scratch by mixing all the data~(also known as joint training) or sequential training for each new incoming dataset may improve the performance on the unseen domains, shown in Figure~\ref{figures: introduction}(b) and (c). Nevertheless, both paradigms still have some limitations respectively. The joint training strategy~\cite{ma2021towards,yan2021towards} requires storing all training data from previous domains when the newly available one arrives, leading to lengthy training time and high storage overhead. Meanwhile, the sequential training strategy will dramatically deteriorate the model's performance among previous domains after training the new domain data, \ie, catastrophic forgetting.

To overcome the aforementioned forgetting, generalization, and storage overhead issues, inspired by the learning mechanism of mammals, we investigate a new task of crowd counting in this paper, termed Lifelong Crowd Counting, which can sustainably learn with the new domain data and concurrently alleviate the catastrophic forgetting and performance drop among preceding domains under the domain-incremental training settings, as shown in Figure~\ref{figures: introduction}(d). One thing to be noted is that the goal of the proposed lifelong crowd counting task is different from that of previous cross-domain and multi-domain crowd counting tasks~\cite{chen2021variational, yan2021towards, ma2021towards}. During the whole lifelong learning process with incremental training data, the goal is to maximize the overall performance among all domains---previously trained, newly arrived, and unseen---instead of only focusing on the target domain performance. We consider the trade-off between the forgetting degree and the generalization ability of the models. In particular, we develop a novel benchmark of domain-incremental lifelong crowd counting with the help of knowledge self-distillation techniques. The proposed benchmark has both \emph{strong generalization ability} on unseen domains and \emph{low forgetting degrees} among seen domains. This helps the model have the sustainable counting capability when new data arrive in the future. In our experiments, we utilize four fruitful crowd counting backbones, CSRNet~\cite{li2018csrnet}, SFANet~\cite{zhu2019dual}, DM-Count~\cite{wang2020distribution}, and DKPNet~\cite{chen2021variational} to illustrate the effectiveness and superiority of our proposed framework. 

The contributions of this work can be summarized as follows.
\begin{itemize}
\item To the best of our knowledge, this is the first work to investigate lifelong crowd counting by considering the catastrophic forgetting and generalization ability issues. Our method may serve as a benchmark for further research in the lifelong crowd counting community. 

\item We design a \textbf{B}alanced \textbf{D}omain \textbf{F}orgetting loss function (BDFLoss) to prevent the model from dramatically forgetting the previous knowledge when being trained on the newly arrived crowd counting dataset.
\item We propose a new quantitative metric \textbf{n}ormalized \textbf{B}ack\textbf{w}ard \textbf{T}ransfer (nBwT) of lifelong crowd counting, to measure the forgetting degree of trained models among seen data domains. 
We treat the Mean Absolute Error (MAE) as the criteria for evaluating model generalization on the unseen data domain.
\item Extensive experiments indicate that our proposed method has a lower degree of forgetting compared with sequential training, and outperforms the joint training strategy on the unseen domain with a much lower MAE score and time and space complexity.
\end{itemize}

\begin{table*}[ht]
\caption{The MAE (mean absolute error) scores of our reproduced DM-Count~\cite{wang2020distribution} model separately trained in a single dataset and tested over other datasets. It shows the obvious performance drop due to the domain discrepancy.}
\centering
\resizebox{0.4\linewidth}{!}{
\begin{tabular}{l|c|c|c|c}
\shline
\textbf{\diagbox{Train}{Test}} & \textbf{SHA} & \textbf{SHB} & \textbf{QNRF} & \textbf{NWPU} \\ \hline
SHA               & \textbf{59.7} & 19.0          & 143.3           & 161.1           \\
SHB               & 124.6          & \textbf{7.0} & 209.9           & 179.1           \\
QNRF              & 69.6          & 14.0          & \textbf{85.6}  & 124.8           \\
NWPU              & 74.7          & 11.7          & 100.9           & \textbf{88.4}  \\ \shline
\end{tabular}
}
\label{results: singledomain}
\end{table*}

\section{Related Work}
\subsection{Crowd counting}
Traditional detection-based and regression-based methods extract the handcrafted features such as SIFT~\cite{lowe1999object} and HoG~\cite{dalal2005histograms} to detect the individual heads~\cite{dalal2005histograms, leibe2005pedestrian, tuzel2008pedestrian, dollar2011pedestrian} or directly regress the count number~\cite{chan2009bayesian}. Nevertheless, these models cannot learn the spatial information of people distribution to make accurate predictions in highly congested scenes. Most of the latest crowd counting approaches are built upon deep learning methods to estimate a density map for a given image. Many researchers design various architectures like fully convolutional networks~\cite{zhang2015cross,wang2015deep}, multi-column networks~\cite{zhang2016single,babu2017switching,sindagi2017generating,boominathan2016crowdnet}, scale aggregation or scale pyramid networks~\cite{jiang2019learning, cao2018scale, chen2019scale,song2021choose,liu2019crowd, zhao2019scale}, and attention mechanisms~\cite{jiang2020attention,guo2019dadnet,sindagi2019ha,liu2019adcrowdnet,zhu2019dual} to extract the multi-scale feature representations to resolve scale variation and non-uniform distribution issues. CSRNet~\cite{li2018csrnet} points out the multi-scale feature redundancies among multi-branch architectures and proposes a new deeper single-column convolutional neural networks~(CNN) with dilated convolutions to capture different receptive fields. ADCNet~\cite{bai2020adaptive} extends the discrete dilated ratio (integer value) into a continuous value to match the large scale variation and self-correct the density map using the Expectation-Maximization~(EM) algorithm. Local region modeling methods~\cite{liu2019counting, jiang2019mask} also help correct the local information. Most off-the-shelf crowd counting models focus on single domain learning. The models will be retrained when the new domain data arrive. In our study, we focus on using a single model to handle multiple incremental datasets for crowd counting.

\subsection{Cross-domain / multi-domain learning}
Many researchers exploit the cross-domain problems~\cite{wu2021dynamic,zou2021coarse,liu2022leveraging,wang2021neuron,wang2019learning} in crowd counting, including cross-scene~\cite{zhang2015cross}, cross-view~\cite{zhang2021cross}, and cross-modal~\cite{liu2021cross}. The Adversarial Scoring Network~\cite{zou2021coarse} is applied to adapt to the target domain from coarse to fine granularity. In addition, cross-domain features can be extracted by the message-passing mechanisms based on a graph neural network~\cite{luo2020hybrid}. A semantic extractor~\cite{han2020focus} has been designed to capture the semantic consistency between the source domain and target domain to enhance the adapted model. A large synthetic dataset (GCC)~\cite{wang2019learning} has been released to study the transferability from synthetic data to real-world data. Quite a few researchers~\cite{yang2020reverse, xiong2019open, shi2019counting} also investigated similar tasks like vehicle counting based on the same architectures from crowd counting. Learning with multiple domains simultaneously~\cite{ma2021towards, yan2021towards, chen2021variational} has also been preliminarily explored, and is required to mix all the data for training at the same time. DCANet~\cite{yan2021towards} proposed a channel attention-guided multi-dilation module to assist the model in learning a domain-invariant representation while DKPNet~\cite{chen2021variational} propagated the domain-specific knowledge with the help of variational attention techniques. Ma~\emph{et al.}~\cite{ma2021towards} developed a scale alignment component to learn an adaptive rescaling factor for each image patch for better crowd counting. In reality, such cross-domain approaches needed a careful alignment module design and placed more emphasis on the target domain performance only, while the multi-domain learning methods required more storage overheads to save old domain data. These methods often achieve limited performance in previous~(source) domains. In contrast, our proposed lifelong crowd counting task is based on training the domains incrementally~(one by one) using a single model, alleviating catastrophic performance drop of the previous domains~(forget less), and maintaining the overall performance in all domains~(count better). The lifelong crowd counting system can mimic the biological brain to learn sustainably in its lifetime inspired by the learning mechanisms of mammals, i.e., integrating the new knowledge increasingly while maintaining previous memories.

\subsection{Lifelong learning}
Lifelong learning attempts to alleviate the catastrophic forgetting issues and enhance the model generalization ability when a system increasingly faces non-stationary data. The mainstream strategies are applied to image classification~\cite{kirkpatrick2017overcoming, rebuffi2017icarl, li2017learning, lopez2017gradient, belouadah2019il2m} and numerical prediction tasks~\cite{he2021clear}, which can be categorized into four groups: model-growth approaches~\cite{rusu2016progressive}, rehearsal-based techniques~\cite{lopez2017gradient, rebuffi2017icarl}, regularization~\cite{kirkpatrick2017overcoming, rebuffi2017icarl} and distillation mechanisms~\cite{li2017learning}. Specifically speaking, the model-growth (\eg~PNN~\cite{rusu2016progressive}) and rehearsal-based methods (\eg~GEM~\cite{lopez2017gradient}) require more computational and memory costs because they either instantiate a new network or replay old data when learning the new classes or tasks. LwF~\cite{li2017learning} is a combination of the distillation networks and fine-tuning to boost the overall performance. However, the aforementioned classification-based lifelong learning approaches can not migrate to the crowd counting task directly because counting is an open-set problem~\cite{xiong2019open} by nature whose value ranges from zero to positive infinity in theory. Latent feature representations with general visual knowledge together with high-level semantic information at the output layer play a crucial role in such dense prediction tasks. Therefore, in this paper, we propose a simple yet effective self-distillation loss at both the feature-level and output-level for lifelong crowd counting to alleviate the catastrophic forgetting with a low time and space complexity.

\section{Methodology}
In this section, we will first introduce concrete formalized definitions of typical crowd counting and the proposed lifelong crowd counting. After that, we will describe the details of our proposed domain-incremental self-distillation lifelong crowd counting benchmark including model architectures and the proposed loss function.
\subsection{Problem formulation}
\subsubsection{Typical crowd counting}
The typical crowd counting task can be regarded as a density map regression problem, training and validating in a single domain, as shown in Figure~\ref{figures: introduction}(a). Suppose one dataset $D_{\mathcal{M}} = \langle X_{\mathcal{M}}, Y_{\mathcal{M}}\rangle$ contains $\mathcal{M}$ training images and the corresponding annotations. Then, a binary map $\textbf{B}$ is easy to obtain given the coordinates of pedestrian heads per image, which can be formally defined as follows: 
\begin{equation}
\textbf{B}_{(i,j)} = 
\left\{
     \begin{array}{ll}
     1, & \text{head center } (i,j) \\
     0, & \text{otherwise} \\
     \end{array}
\right.
\end{equation}

The ground truth density map $Y$ is generated by employing the Gaussian kernel $G_\sigma$ to smooth the binary map. 
\begin{equation}
    Y = G_\sigma \circledast \textbf{B}.
\end{equation}
Here, `$\circledast$' represents the convolution operation. Then, the typical crowd counting is transformed to regress the generated density maps. The pixel-level $\mathcal{L}_2$ loss is the most commonly used one to optimize the model $\mathcal{F}(\cdot;\theta)$ with the parameter $\theta$  by minimizing the difference between predictions and ground truths:
\begin{equation}
    \min_\theta \frac{1}{\mathcal{M}}\sum_{m=1}^{\mathcal{M}}\mathcal{L}_2(\mathcal{F}(X_m; \theta), Y_m)\;.
\end{equation}

\subsubsection{Lifelong crowd counting}
We propose a new, challenging yet practical crowd counting task, \ie, lifelong crowd counting, for investigating the catastrophic forgetting and model generalization problems in training domain-incremental datasets. Different from previous works that only maintained good performance in a single target domain, the lifelong crowd counting model could be sustainably optimized over the new incoming datasets to maximize the performance among all domains. 

For convenience, we first define some key notations as follows and introduce the details of the lifelong crowd counting process. A sequence of $\mathcal{N}$ domain datasets $\{\mathcal{D}_1, \mathcal{D}_2,\ldots,\mathcal{D}_{\mathcal{N}}\}$ is prepared to train the lifelong crowd counter $\mathcal{G}^*(\cdot;\psi)$ with parameters $\psi$ one by one. $X_{\mathcal{M}_t}^{(t)}$ and $Y_{\mathcal{M}_t}^{(t)}$ are the training images and corresponding ground truth density maps with $\mathcal{M}_t$ samples from the $t$-th domain $\mathcal{D}_t$, respectively. Here, we assume different datasets are coming from different domains with their own distinct data distributions, {\textit{i.e.}} $p(X^{(i)})\neq p(X^{(j)}),\; i\neq j$ because they are normally captured from different cameras or different scenarios like streets, museums, and gymnasiums. The model is initially trained from scratch over the first domain and then trained and optimized by the rest of the other datasets sequentially. The optimal object $\psi^{*}$ is defined as follows:
\begin{equation}
    \mathop{\arg\min}\limits_{\psi} \sum_{t=1}^{\mathcal{N}} \mathbb{E}_{(X_{\mathcal{M}_t}^{(t)},Y_{\mathcal{M}_t}^{(t)})} [ \mathcal{L} (\mathcal{G}^{(t)}(X_{\mathcal{M}_t}^{(t)};\psi), Y_{\mathcal{M}_t}^{(t)}) ],
\end{equation}
where $\mathcal{G}^{(t)}(\cdot; \psi)$ represents the $t$-th model for training $t$-th dataset $X_{\mathcal{M}_t}^{(t)}$ with $\mathcal{M}_t$ samples. The ultimate model is expected to achieve decent performance among seen and unseen domains. What deserves to be pointed out is that lifelong crowd counting is distinct from cross-domain tasks with different optimization objectives, as well as the training settings. In lifelong crowd counting, the goal is to maximize the performance on both seen and unseen domains instead of maximizing the target domain performance only. Especially when the training data from previous domains are absent or unavailable, lifelong crowd counters could still work efficiently because they are trained and updated only by the newly-arrived domain dataset one after another.

\begin{figure*}[ht]
   \begin{center}
   \includegraphics[width=0.6\linewidth]{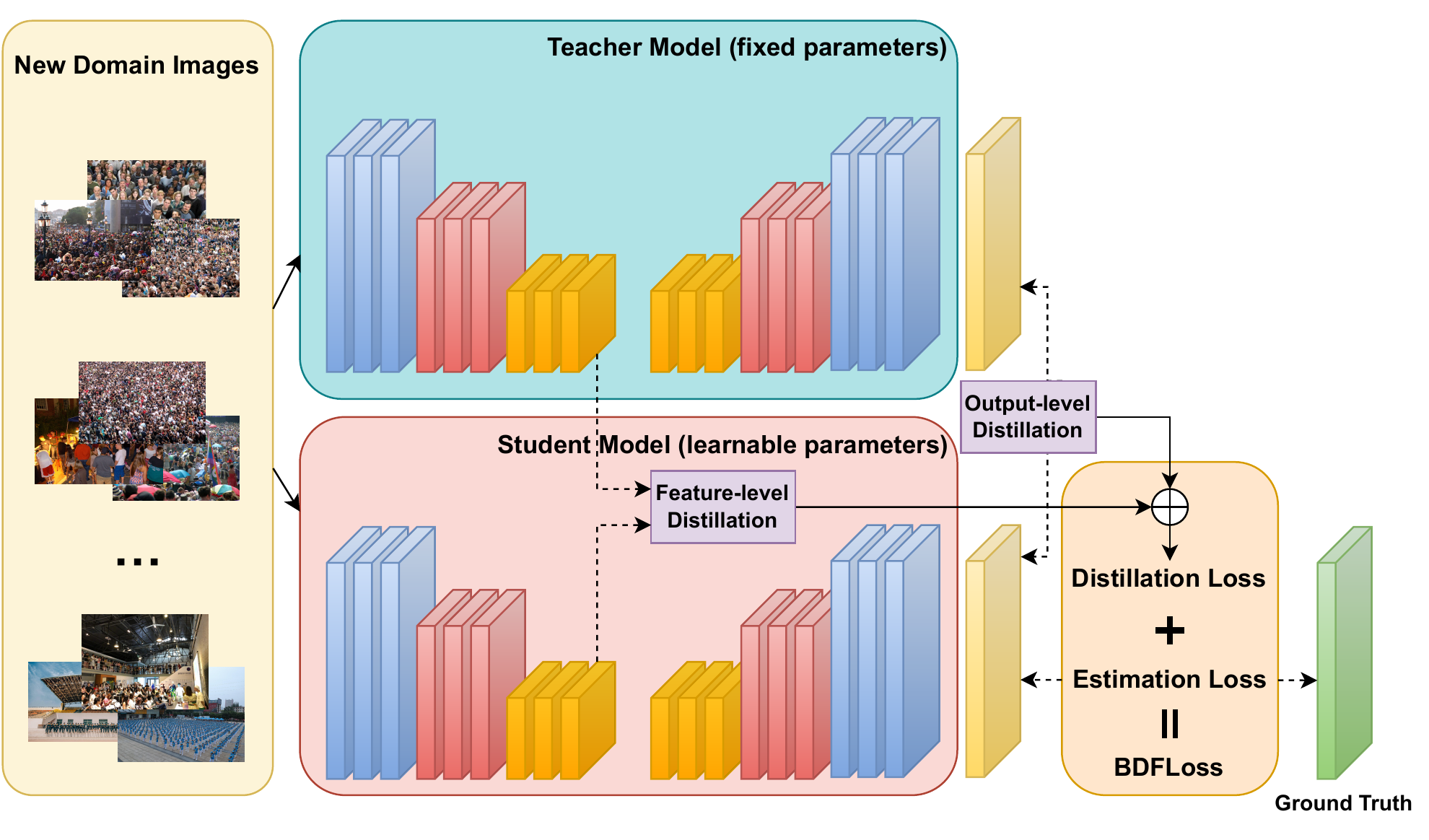}
   \end{center}
   \caption{The overall architecture of our proposed domain-incremental self-distillation learning benchmark~(FLCB).}
   \label{figures: model}
\end{figure*}

\subsection{Overview of our proposed framework}
Our proposed framework focuses on tackling the catastrophic forgetting and generalization issues under the circumstances of domain-incremental training settings. In this paper, we simply regard different crowd counting datasets as different domains because the statistics~(mean and variance) of people counts are different. The detailed explanations of the domain concept can be seen in Appendix~\ref{appendix: domain concept}. To be more specific, we propose a novel domain-incremental self-distillation lifelong crowd counting benchmark for sustainable learning with newly arrived data and without an obvious performance drop among previous domains. The key factor is how to effectively leverage the previously learned meaningful knowledge when training over the data from a new domain for better crowd counting. Inspired by the knowledge distillation technique, we expect to use a well-trained model among old domains~(teacher model) to guide the currently-optimized model with new domain data~(student model) to mitigate performance drop among previous domains, considering that the old data may be unavailable. The overview of our proposed framework is illustrated in Figure~\ref{figures: model}. We design a self-distillation mechanism plugged into both feature-level and output-level layers of the network to constrain the output distribution similarities between the teacher and student models, which can reuse the learned knowledge when facing the new domain data without storing and training the old data repeatedly. More details will be discussed in Section~\ref{section: bdfloss}. The ultimate model is expected to be deployed to an arbitrary domain to estimate the people counts. 

For better understanding, the overall training pipeline is described in detail as shown in Algorithm~\ref{training_pipeline}. A queue $Q$ collects $\mathcal{N}$ increasingly-arrived datasets from different domains to be trained one by one. First, we initialize the first model $\mathcal{G}^{(1)}(\cdot;\psi)$ by training the first available dataset $\mathcal{D}_1$ in the queue $Q$. Another queue $P$ is prepared for future evaluation, which receives the test set popped from $Q$. After that, the model will be trained and optimized by the subsequent datasets from $\mathcal{D}_2$ to $\mathcal{D}_\mathcal{N}$, repeating the following main steps until the queue $Q$ is empty:
\begin{itemize}
\item[(1)] Pop the $t$-th dataset $\mathcal{D}_t$ from the queue $Q$ for training.
\item[(2)] Copy the parameters of the last well-trained model $\mathcal{G}^{(t-1)}$ to the model $\mathcal{F}(\cdot;\theta)$ as a teacher network for distillation.
\item[(3)] Train the current $t$-th model $\mathcal{G}^{(t)}(\cdot;\psi)$ over the $t$-th dataset $\mathcal{D}_{t}$ via the self-distillation loss we proposed.
\item[(4)] Push the $t$-th dataset $\mathcal{D}_{t}$ into the queue $P$ for evaluation when the model converges.
\end{itemize}
Note that the parameters $\theta$ of the model $\mathcal{F}(\cdot;\theta)$ are frozen during the lifelong training process. The fixed model is regarded as a teacher network to guide the current student network $\mathcal{G}^{(t)}(\cdot;\psi)$ with learnable parameters $\psi$ to remember old meaningful knowledge for better crowd counting. Eventually, we obtain the final model with the best parameters $\psi^*$, which can continue to be trained by using our proposed framework when the new labeled data are ready in the future. Because we do need to store any previously-seen training data to be replayed to train our model, the time and space complexity is approximately $O(\mathcal{N})$ and $\Omega(\mathcal{M})$ which is superior to $O(\mathcal{N}^2)$ and $\Omega(\mathcal{N}\times \mathcal{M})$ of joint training. $\mathcal{M}$ is the maximum of  $\mathcal{M}_i$. Although the distillation mechanism is required to save an additional model, its storage overhead is negligible compared to storing the entire dataset for retraining.

\begin{algorithm*}[ht]
    \renewcommand{\algorithmicrequire}{\textbf{Input:}}
    \renewcommand{\algorithmicensure}{\textbf{Output:}}
    \caption{FLCB training pipeline.}
    \label{training_pipeline}
    \textbf{Notations:}\\
        $X^{(t)}_{\mathcal{M}_t}:$ The $t$-th training dataset with $\mathcal{M}_t$ samples.\\
        $Y^{(t)}_{\mathcal{M}_t}:$ The corresponding density maps of $X^{(t)}_{\mathcal{M}_t}$.\\
        $\mathcal{M}_1, \mathcal{M}_2, ..., \mathcal{M}_{\mathcal{N}}:$ Samples of each dataset.\\
        $P:$ A queue containing previously seen datasets.\\
        $Q:$ A queue containing future unseen datasets.\\
        $\mathcal{F}^{(t)}(\cdot;\theta):$ Teacher model with fixed params $\theta$ at $t$-th step.\\
        $\mathcal{G}^{(t)}(\cdot;\psi):$ Student model with updated params $\psi$ at $t$-th step. 
    \begin{algorithmic}[1]
        \REQUIRE $\{\mathcal{D}_1,\mathcal{D}_2,...,\mathcal{D_N}\}$: A sequence of $\mathcal{N}$ domain datasets, $\mathcal{D}_i = \langle X^{(i)}_{\mathcal{M}_i}, Y^{(i)}_{\mathcal{M}_i} \rangle $.
        \ENSURE The optimal model parameter $\psi^*$.
        \STATE $P \gets \varnothing $;
        \STATE $Q \gets \{\mathcal{D}_1, \mathcal{D}_2, \ldots, \mathcal{D}_{\mathcal{N}}\}$;
        \STATE $\langle X^{(1)}_{\mathcal{M}_1},Y^{(1)}_{\mathcal{M}_1}\rangle \gets Q.top()$;
        \STATE $Q.pop()$;
        \STATE \textbf{Train} $\mathcal{G}^{(1)}(X^{(1)}_{\mathcal{M}_1};\psi)$\;
        \STATE $\psi^* \gets \mathop{\arg\min}_{\psi} \mathcal{L}^{(1)}_{\text{count}}(\cdot;\psi)$\;
        \STATE $P.push(X^{(1)})$\;
        \FOR {$t=2,3,\ldots,\mathcal{N}$}   
            \STATE $\mathcal{F}^{(t-1)}(\cdot;\theta) \gets \mathcal{G}^{(t)}(\cdot;\psi^*)$\;
            \STATE $ \langle X^{(t)}_{\mathcal{M}_t},Y^{(t)}_{\mathcal{M}_t} \rangle \gets Q.top() $\;
            \STATE \textbf{Train} $\mathcal{G}^{(t)}(X^{(t)}_{\mathcal{M}_t};\psi)$\;
            \STATE $\psi^* \gets \mathop{\arg\min}_{\psi} \mathcal{L}^{(t)}_{\text{count}}(\cdot;\psi) + \lambda \mathcal{L}^{(t)}_{\text{distill}}(\cdot;\theta,\psi) $\;
            \STATE $P.push(\langle X^{(t)}_{\mathcal{M}_t},Y^{(t)}_{\mathcal{M}_t}\rangle)$\;
            \STATE $Q.pop()$\;
            \STATE \textbf{Test} all seen datasets in $P$ with $\mathcal{G}^{(t)}(\cdot;\psi^*)$\;
        \ENDFOR
        
        \STATE \textbf{Return} $\psi^*$\;
    \end{algorithmic}  
    // Time Complexity: $O(\mathcal{N})$\\
    // Space Complexity: $\Omega(\mathcal{M})$\\
    // $\mathcal{M}=\max\{\mathcal{M}_i| i = 1,2,...,\mathcal{N}\}$
\end{algorithm*}

\subsection{Balanced domain forgetting loss}\label{section: bdfloss}
To balance the model plasticity (the ability to learn new data) and stability (the ability to remember previous knowledge), we propose a novel balanced domain forgetting loss function, {\textit{i.e.}}, BDFLoss, mainly consisting of counting loss and self-distillation loss. We integrate the optimal transport loss in our basic $\mathcal{L}_1$ counting loss in this study because it has tighter generalization error bounds~\cite{wang2020distribution}. $\mathcal{L}_1$ counting loss is defined as follows:
\begin{equation}
    \mathcal{L}_1(\text{Y}, \hat{\text{Y}}) = \frac{1}{\mathcal{M}} \sum_{i=1}^{\mathcal{M}}|\text{Y}_i - \hat{\text{Y}}_i|,
\end{equation}
where $\mathcal{L}_1(\cdot, \cdot)$ loss computes the difference between the predictions and actual counts.

The optimal transport loss $\mathcal{L}_{\text{OT}}$ is used to minimize the distribution discrepancy between the predicted density maps and the point-annotated binary maps, and is defined as follows:
\begin{equation}
\begin{split}
    \mathcal{L}_{\text{OT}}(\text{Y}, \hat{\text{Y}}) &= \mathcal{W}_c\left(\tfrac{\text{Y}}{||\text{Y}||_1}, \tfrac{\hat{\text{Y}}}{||\hat{\text{Y}}||_1}; \mathcal{C}\right) \\
    &= \left \langle \alpha^{*}, \tfrac{\text{Y}}{||\text{Y}||_1} \right\rangle + \left \langle \beta^{*}, \tfrac{\hat{\text{Y}}}{||\hat{\text{Y}}||_1} \right\rangle ,
\end{split}
\end{equation}
where $\mathcal{W}_{c}(\mu, v;\mathcal{C})$ is the optimal transport loss with the transport cost $\mathcal{C}$. It aims at minimizing the cost to transform one probability distribution $\mu$ to another $v$. $\mathcal{C}$ is defined as the quadratic transport cost here. $\alpha^{*}$ and $\beta^{*}$ are the optimal solutions of its dual problem as follows:
\begin{equation}
    \max_{\alpha, \beta} \left \langle \alpha, \mu \right \rangle + \left \langle \beta, v \right \rangle \quad \text{s.t.} \quad \alpha_i + \beta_j \leq \mathcal{C}_{ij}, \forall i,j.
\end{equation}

To improve the approximation of the low-density regions of images, we also embed a normalized regularization item $\mathcal{L}_{\text{r}}$, defined as follows:
\begin{equation}
    \mathcal{L}_{\text{r}}(\text{Y}, \hat{\text{Y}}) =  \frac{1}{\mathcal{M}} \sum_{i=1}^{\mathcal{M}}\frac{1}{2}\left\|\tfrac{\text{Y}_i}{||\text{Y}_i||_1}-\tfrac{\hat{\text{Y}}_i}{||\hat{\text{Y}}_i||_1}\right\|_1.
\end{equation}

Thus, the total count loss is made up of the three aforementioned loss functions with two hyper-parameters, $\eta$ and $\gamma$, which are set to 0.1 and 0.01, respectively, in our experiments.
\begin{equation}
\mathcal{L}_\text{count}=\mathcal{L}_1 + \eta \mathcal{L}_{\text{OT}} + \gamma \mathcal{L}_{\text{r}}.
\end{equation}

When training to the $t$-th domain, the performance among previous domains may degrade dramatically, {\textit{i.e.}, catastrophic forgetting}, if no constraints are imposed. The self-distillation loss $\mathcal{L}_{\text{distill}}$ is designed to help the model forget less and count better during the lifelong learning process. To be more specific, we regard the current training model $\mathcal{G}^{(t)}(\cdot)$ as the student model, which can be guided by the teacher model $\mathcal{G}^{(t-1)}(\cdot)$ well-trained at the previous step, as shown in Figure~\ref{figures: model}. The student model is not expected to forget some previously-learned knowledge when training in the new domain. Normally, the deep layers of a CNN with a large receptive field contain task-specific and high-level semantic information while the intermediate layers include general visual knowledge. They are mutually beneficial and complementary, and assist the model in remembering the helpful knowledge learned previously, during the lifelong crowd counting process. Thus, we deploy the self-distillation loss at both the feature level and the output level when the $t$-th new domain dataset arrives for training.
\begin{align}
    \mathcal{L}^{(t)}_{\text{distill}} =& \frac{1}{\mathcal{M}_t}\sum_{i=1}^{\mathcal{M}_t} \Big( \underbrace{||\mathcal{G}^{(t-1)}(X^{(t)}_i) - \mathcal{G}^{(t)}(X^{(t)}_i)||^2}_{\text{output-level distillation}} \notag\\&+  \underbrace{||\mathcal{H}^{(t-1)}(X^{(t)}_i) - \mathcal{H}^{(t)}(X^{(t)}_i) ||^2}_{\text{feature-level distillation}} \Big),
\end{align}
where $\mathcal{H(\cdot)}$ denotes the feature extractor of the model $\mathcal{G}(\cdot)$. Since the similarity metric is not our crucial research point in this paper, we just choose the $L_2$ loss for simplicity. To sum up, the BDFLoss is made up of these two components within the hyper-parameter $\lambda$.
\begin{equation}
   \mathcal{L}_\text{BDF} = \mathcal{L}_\text{count} + \lambda \mathcal{L}_\text{distill},
\end{equation}
where the hyper-parameter $\lambda$ is applicable as a trade-off between model plasticity and stability. It is the same as vanilla sequential fine-tuning when $\lambda$ is equal to 0.

\subsection{Model architectures}
Our proposed domain-incremental self-distillation lifelong crowd counting benchmark is model-agnostic. Therefore, to illustrate its effectiveness, we integrate it into several state-of-the-art crowd counting backbone models without the bells and whistles, CSRNet~\cite{li2018csrnet}, SFANet~\cite{zhu2019dual}, DM-Count~\cite{wang2020distribution}, and DKPNet~\cite{chen2021variational}. Because the attention map supervision of SFANet may introduce some biases in the experimental comparisons and the source code of DKPNet is not released, we make the following modifications in our experiments. A small improvement of SFANet is that we enable the network to learn the attention map adaptively based on training images without generating additional attention maps for supervision. We modify the DKPNet-baseline in our experiments because we only focus on investigating the effectiveness of our proposed framework on forgetting and generalization under different model capacities.

\section{Experiment Settings}
In this section, we will briefly introduce four datasets that we used in our experiments, the training settings, and some hyper-parameter selections.

\subsection{Datasets}
We train and evaluate our model in the public crowd counting datasets, {\textit{i.e.}}, ShanghaiTech PartA~\cite{zhang2016single}, ShanghaiTech PartB~\cite{zhang2016single}, UCF-QNRF~\cite{idrees2018composition}, NWPU-Crowd~\cite{wang2020nwpu}, and JHU-Crowd++~\cite{sindagi2019jhucrowd}, as shown in Table \ref{table: datasets}. To illustrate the generalization of different training paradigms, we have to select one of them as the unseen dataset that could never be trained during the domain-incremental lifelong learning process. In our experiments, we take the JHU-Crowd++ dataset as an unseen one because it has a variety of diverse scenarios and unconstrained environmental conditions~\cite{sindagi2019jhucrowd}. The synthetic dataset GCC~\cite{wang2019learning} is also utilized to analyze the synthetic-to-real generalization performance under the lifelong crowd counting settings.

\begin{table*}[ht]
\caption{The number of images used to train models on different datasets. Last three columns illustrate the minimum, maximum, and average number of people per image.}
\centering
\resizebox{0.6\linewidth}{!}{
\begin{tabular}{l|c|c|c|c|c}
\shline
\textbf{Dataset}            & \tabincell{c}{\textbf{Raw samples}/\textbf{Training samples}} & \textbf{Test}& \textbf{Min}& \textbf{Max} & \textbf{Avg} \\ \hline
ShanghaiTech PartA & 300/300 & 182 &33 &3,139 & 501\\
ShanghaiTech PartB & 400/400 & 316 &9 &578 &123\\
UCF-QNRF           & 1,201/1,201 & 334 &49 &12,865 & 815     \\
NWPU-Crowd         & 3,609/3,609 & 1,500 &0 &20,033 &418     \\
JHU-Crowd++          & 2,772/0 & 1,600 &0 &25,791 &346  \\\hline
GCC &\multicolumn{2}{c|}{15,212} & 0 & 3,995 & 501 \\
\shline
\end{tabular}
}
\label{table: datasets}
\end{table*}

\subsection{Implementation details}
We strictly follow the same basic image preprocessing settings in most recent literature~\cite{wang2020distribution, ma2019bayesian, li2018csrnet, zhu2019dual}. The crop size is $256\times 256$ for SHA, and $512\times 512$ for SHB, QNRF, and NWPU datasets. To generate the density map as ground truth, we just adopt the fixed Gaussian kernel whose variance $\sigma$ is set to 15 for all datasets. Several useful augmentations like random horizontal flipping with a probability of $0.5$ and normalization are also applied to those images before training. The hyper-parameter $\lambda$ in the loss function is set to 0.5 to be a trade-off between model plasticity and stability. We use the fixed learning rate of 1e-5, a simple weight decay of 5e-4, and an Adam optimizer in all of our experiments. We use the Pytorch framework and NVIDIA GeForce RTX 3090 GPU workstation to conduct our experiments.

\subsection{Evaluation metrics}
The catastrophic forgetting phenomenon often exists in domain-incremental learning. To evaluate how much on earth the model forgets in the previous domains and make a fair comparison with other methods, we propose a new metric, called \textbf{n}ormalized \textbf{B}ack\textbf{w}ard \textbf{T}ransfer (nBwT). With the help of nBwT, the total forgetfulness over $t$ incremental domains could be measured to determine whether the model is equipped with the sustainable learning ability. The normalization operation we introduced in nBwT could eliminate the potential negative impact because of the different learning difficulties in different domains.
\begin{equation}
    \text{nBwT}_t = \frac{1}{t-1} \sum_{i=1}^{t-1}  \frac{e_{t,i} - e_{i,i}}{e_{i,i}}\;,\; t =2,...,\mathcal{N},
\end{equation}
where $e_{t,i}$ is the test MAE score of $i$-th dataset when obtaining the optimal model on the $t$-th dataset, and $i<t$. $\text{nBwT}_t$ is the accumulation of the forgetting performance among all previous $t-1$ domain datasets. The non-zero divisor $e_{i,i}$ is a normalization factor. The larger the nBwT value is, the greater the model forgetting degree is. A value that is <0 indicates that the model has attained a positive performance improvement among previously trained datasets. The theoretical lower bound of $\text{nBwT}_t$ is $-\frac{1}{t-1}$ when $e_{t,i}$ equals zero.  

Furthermore, we propose two reasonable and impartial criteria, {\textit{i.e.}}, mMAE and mRMSE~(the respective means of MAE and RMSE in $\mathcal{N}$ datasets)
to evaluate the roughly overall counting precision of the lifelong crowd counting task as follows.
\begin{align}
   \text{mMAE}=\frac{1}{\mathcal{N}} \sum_{i=1}^\mathcal{N} \frac{1}{\mathcal{M}_i} \sum_{j=1}^{\mathcal{M}_i}  | \hat{Y_j} - Y_j | ,
   \end{align}
\begin{align}
   \text{mRMSE}=\frac{1}{\mathcal{N}} \sum_{i=1}^\mathcal{N} \sqrt{\frac{1}{\mathcal{M}_i} \sum_{j=1}^{\mathcal{M}_i} || \hat{Y_j} - Y_j ||^2},
\end{align}
where $\mathcal{M}_i$ denotes the number of images from the $i$-th test set. $\hat{Y_j}$ and $Y_j$
are the predicted counts and actual counts of the $j$-th image, respectively. The mMAE and mRMSE reduce to standard MAE and RMSE when $\mathcal{N}$ is equal to 1.

In addition, we still use the standard MAE score on the unseen JHU-Crowd++ dataset to compare the model generalization within different training strategies.

\begin{table*}[ht]
\caption{The results with different domain-incremental lifelong learning methods. `*' represents our reproduced results of modified approaches.}
\resizebox{\linewidth}{!}{
\begin{tabular}{l|cc|cc|cc|cc|cc|cc}
\shline
\multirow{2}{*}{\textbf{Model}} & \multicolumn{2}{c|}{\textbf{SHA}}                   & \multicolumn{2}{c|}{\textbf{QNRF}}                  & \multicolumn{2}{c|}{\textbf{SHB}}                   & \multicolumn{2}{c|}{\textbf{NWPU}}                  & \multirow{2}{*}{\textbf{mMAE}} & \multirow{2}{*}{\textbf{mRMSE}} & \multicolumn{2}{c}{\textbf{JHU (unseen)}}          \\\cline{2-9}\cline{12-13}
& MAE & RMSE & MAE & RMSE & MAE & RMSE & MAE & RMSE &  &  & MAE & RMSE \\\hline
LwF*~\cite{li2017learning}                            & 62.3                    & 104.4                    & 81.4                    & 133.4                    & 11.5                    & 18.2                     & 90.8                    & 395.2                    & 61.5                                               & 162.8                                               & 90.4                    & 298.2                    \\
EwC*~\cite{kirkpatrick2017overcoming}                            & 64.9                    & 117.2                    & 88.5                    & 171.7                    & 10.2                    & 17.6                     & 84.2                    & 377.7                    & 62.0                                               & 171.1                                               & 85.9                    & 294.1                    \\
FLCB~(Ours)                            & 68.8                    & 113.9                    & 84.3                    & 160.1                    & 7.8                     & 12.2                     & 76.6                    & 364.2                    & \textbf{59.4}                                      & \textbf{162.6}                                      & \textbf{84.8}           & \textbf{264.8}     \\
\shline
\end{tabular}
}\label{table: llcomparison}
\end{table*}

\begin{table*}[ht]
\caption{The quantitative results with different paradigms to compare the forgetting degree and overall performance. We take the sequential training as our BASELINE and take the joint training as JOINT for reference. FLCB is our proposed method.}
\label{table: catastrophic forgetting}

\resizebox{\linewidth}{!}{
\begin{tabular}{l|l|cc|cc|cc|cc|c|c|c|c|c}
\shline
\multirow{2}{*}{\textbf{Model}} &
  \multicolumn{1}{c|}{\multirow{2}{*}{\textbf{Method}}} &
  \multicolumn{2}{c|}{\textbf{SHA}} &
  \multicolumn{2}{c|}{\textbf{QNRF}} &
  \multicolumn{2}{c|}{\textbf{SHB}} &
  \multicolumn{2}{c|}{\textbf{NWPU}} &
  \multicolumn{1}{c|}{\multirow{2}{*}{\textbf{mMAE}}} &
  \multicolumn{1}{c|}{\multirow{2}{*}{\textbf{mRMSE}}} &
  \multicolumn{1}{c|}{\multirow{2}{*}{\textbf{nBwT}}}&
  \multicolumn{1}{c|}{\multirow{2}{*}{\textbf{\# params.}}}&
  \multicolumn{1}{c}{\multirow{2}{*}{\textbf{MACs}}} \\ \cline{3-10}
 &
  \multicolumn{1}{c|}{} &
  \textbf{MAE} &
  \textbf{RMSE} &
  \textbf{MAE} &
  \textbf{RMSE} &
  \textbf{MAE} &
  \textbf{RMSE} &
  \textbf{MAE} &
  \textbf{RMSE} &
  \multicolumn{1}{c|}{} &
  \multicolumn{1}{c|}{} &
  \multicolumn{1}{c|}{} &
  \multicolumn{1}{c|}{} &
  \multicolumn{1}{c}{} \\ \hline
\multirow{4}{*}{\tabincell{c}{CSRNet\\\cite{li2018csrnet}}} & BASELINE & 98.4 & 168.1 & 123.9 & 225.3 & 13.4 & 19.1 & 114.5 & 456.5 & 87.6 & 217.3 & 0.424 & \multirow{4}{*}{16.26M} & \multirow{4}{*}{27.07G} \\
                        & LwF*          & 71.5 & 122.4 & 107.4 & 198.9 & 11.3 & 16.7 & 123.3 & 520.3 & \textbf{78.4} & 214.6 &-0.042 &                       \\
                        & FLCB         & 66.6 & 100.4 & 112.5 & 198.6 & 13.0 & 22.0 & 121.4 & 473.2 & \textbf{78.4} & \textbf{198.6} &\textbf{-0.102} &                       \\
                        & JOINT & 64.0 & 100.6 & 109.0 & 199.7 & 14.0 & 18.6 & 124.8 & 499.4 & \textit{78.0} & \textit{204.6} & - & \\\hline
\multirow{4}{*}{\tabincell{c}{SFANet\\\cite{zhu2019dual}}} & BASELINE & 85.4 & 141.3 & 112.6 & 200.7 & 14.8 & 18.1 & 106.9 & 463.7 & 79.9 & 206.0 & 0.545 & \multirow{4}{*}{17.02M}& \multirow{4}{*}{27.28G} \\
                        & LwF*          & 75.0 & 128.5 & 101.3 & 177.2 & 11.5 & 19.0 & 108.3 & 450.0 & 74.0 & 193.7 &-0.002 &                       \\
                        & FLCB         & 69.4 & 110.9 & 103.7 & 176.6 & 12.7 & 20.9 & 108.8 & 445.0 & \textbf{73.7} & \textbf{188.4} &\textbf{-0.097} &                       \\
                        & JOINT & 77.7 & 124.0 & 136.8 & 236.3 &14.0 &17.3 & 127.8 & 458.5 & \textit{89.1} & \textit{209.0} & - &\\\hline
\multirow{4}{*}{\tabincell{c}{DM-Count\\\cite{wang2020distribution}}}  & BASELINE & 76.0 & 122.2 & 94.1 & 154.1 & 9.6 & 17.5 & 108.3 & 481.4 & 72.0 & 193.8 & 0.176 & \multirow{4}{*}{21.50M}& \multirow{4}{*}{26.99G} \\
                        & LwF*          & 74.6 & 124.1 & 90.2 & 164.9 & 9.4 & 14.9 & 86.9 & 375.4 & 65.3 & 169.8 
                        &  0.049    &    \\
                        & FLCB         & 69.2 & 113.2 & 95.4 & 166.0 & 9.7 & 15.6 & 83.6 & 370.8 & \textbf{64.5} & \textbf{166.4} &    \textbf{-0.013} &                \\
                        & JOINT & 78.2 & 129.3 & 86.7 & 153.3 &7.9 &13.0 & 88.5 & 393.8 & \textit{65.3} & \textit{172.4} & - &\\
                        \hline
\multirow{4}{*}{\tabincell{c}{DKPNet\\\cite{chen2021variational}}}  & BASELINE & 92.9 & 157.8 & 100.1 & 179.4 & 7.7 & 12.4 & 90.0 & 393.6 & 72.7 & 185.8 & 0.371 & \multirow{4}{*}{13.28M}& \multirow{4}{*}{10.38G} \\
                        & LwF*          & 62.3 & 104.4 & 81.4 & 133.4 & 11.5 & 18.2 & 90.8 & 395.2 & 61.5 & 162.8 
                        &    -0.009    &        \\
                        & FLCB         & 68.8 & 113.9 & 84.3 & 160.1 & 7.8 & 12.2 & 76.6 & 364.2 & \textbf{59.4} & \textbf{162.6} &  \textbf{-0.010}  &              \\ 
                        & JOINT & 65.0 & 108.5 & 86.0 & 163.3 &8.4 &13.2 & 81.2 & 357.7 & \textit{60.2} & \textit{160.7} & - &\\
\shline
\end{tabular}}
\end{table*}

\section{Experimental Results}
In this section, we first evaluate the overall performance and the generalization ability of our proposed FLCB framework compared with those of two classical continual learning approaches~\cite{li2017learning, kirkpatrick2017overcoming} in Table~\ref{table: llcomparison}. Then, we demonstrate the difference between our FLCB and the other three learning strategies, especially for analyzing their respective forgetting degrees among the trained datasets (SHA, SHB, QNRF, and NWPU), and their generalization abilities on the unseen dataset (JHU-Crowd++). The synthetic-to-real experiments are also conducted considering the data privacy issues and some ethical policies.
\begin{table*}[ht]
\caption{The forgetting performance in the intermediate process of lifelong crowd counting among four models with FLCB. The data underlined are all less than zero, which has a positive effect on the overall performance of among past domains.}

\resizebox{\linewidth}{!}{
\begin{tabular}{l|c|cc|cc|cc|cc|c|c|c}
\shline
        \multirow{2}{*}{\textbf{Method~(FLCB)}}    & \multirow{2}{*}{\textbf{Model}} & \multicolumn{2}{c|}{\textbf{SHA}} & \multicolumn{2}{c|}{\textbf{QNRF}} & \multicolumn{2}{c|}{\textbf{SHB}} & \multicolumn{2}{c|}{\textbf{NWPU}} & \multirow{2}{*}{\textbf{mMAE}} & \multirow{2}{*}{\textbf{mRMSE}} &\multirow{2}{*}{\textbf{nBwT}}\\ \cline{3-10}
        & &\textbf{MAE} & \textbf{RMSE} &\textbf{MAE} & \textbf{RMSE} &\textbf{MAE} & \textbf{RMSE} &\textbf{MAE} & \textbf{RMSE} &  & &\\ \hline
SHA$\rightarrow$QNRF    & CSRNet & 73.9    & 111.7    & 121.8     & 225.3 &-&-&-&-&97.9&168.5 & 0.068 \\
SHA$\rightarrow$QNRF    & SFANet & {73.4} & {114.4} & 111.3 & 200.4 &-&-&-&-& 92.4 &157.4 & {0.225}   \\
SHA$\rightarrow$QNRF & DM-Count & {65.2} & {117.2} & 84.8 &149.0 & - & - & - & - &75.0 &133.1 & {\textbf{0.058}} \\
SHA$\rightarrow$QNRF & DKPNet & 62.1 & {103.9} & 82.9 & 149.7 & - & - & - & - &\textbf{72.5} & \textbf{126.8}& 0.078 \\\hline

SHA$\rightarrow$QNRF$\rightarrow$SHB    & CSRNet  &73.9 & 111.7 & 121.8 & 225.3 & 16.1 & 29.9&-&-&70.6 &122.3 &0.034 \\
SHA$\rightarrow$QNRF$\rightarrow$SHB    & SFANet  & {73.4} & {114.4} & {111.3} & {200.4} & 20.5 & 31.5 &-&-&68.4 & 115.4&  {0.113}   \\
SHA$\rightarrow$QNRF$\rightarrow$SHB & DM-Count &  {65.2} & {117.2} & {84.8} & {149.0} & 13.6 & 25.6 & - & - &54.3 & 97.3& {0.029} \\
SHA$\rightarrow$QNRF$\rightarrow$SHB & DKPNet & {63.5} & {109.6} & {86.4} & {147.5} & 10.3 & 17.3 & - & - &\textbf{53.4} &\textbf{91.5} & \underline{\textbf{-0.014}} \\ \hline

SHA$\rightarrow$QNRF$\rightarrow$SHB$\rightarrow$NWPU & CSRNet &66.6 &{100.4} &{112.5} & {198.6} &{13.0} &22.0 &121.4 &473.2&78.4 & 198.6&\underline{\textbf{-0.102}}     \\ 
SHA$\rightarrow$QNRF$\rightarrow$SHB$\rightarrow$NWPU & SFANet &{69.4} &{110.9} &{103.7} &{176.6} &{12.7} &20.9 & 108.8 & {445.0}&73.7 &188.4 & \underline{-0.097}    \\  
SHA$\rightarrow$QNRF$\rightarrow$SHB$\rightarrow$NWPU &
DM-Count & {69.2} &{113.2} &95.4 &166.0 &9.7 &{15.6} &{83.6} &{370.8} &64.5 &166.4 &\underline{-0.013} \\
SHA$\rightarrow$QNRF$\rightarrow$SHB$\rightarrow$NWPU & DKPNet &{68.8} &{113.9} &{84.3} &{160.1} &7.8 &{12.2} &{76.6} &{364.2} &\textbf{59.4} &\textbf{162.6} &\underline{-0.010} \\
\shline
\end{tabular}}
\label{results: length}
\end{table*}

\begin{table*}[ht]
\caption{The forgetting degree comparison results with different hyper-parameters $\lambda$.}
\label{results: lambda}
\resizebox{\linewidth}{!}{
\begin{tabular}{l|c|c|cc|cc|cc|cc|c}
\shline
\multirow{2}{*}{\textbf{Method~(FLCB)}} &
  \multirow{2}{*}{\textbf{Model}} &
  \multirow{2}{*}{\textbf{$\lambda$}} &
  \multicolumn{2}{c|}{\textbf{SHA}} &
  \multicolumn{2}{c|}{\textbf{QNRF}} &
  \multicolumn{2}{c|}{\textbf{SHB}} &
  \multicolumn{2}{c|}{\textbf{NWPU}} &
  \multirow{2}{*}{\textbf{nBwT}} \\\cline{4-11}
 &
   &
   &
  \textbf{MAE} &
  \textbf{RMSE} &
  \textbf{MAE} &
  \textbf{RMSE} &
  \textbf{MAE} &
  \textbf{RMSE} &
  \textbf{MAE} &
  \textbf{RMSE} &
   \\\hline
SHA$\rightarrow$QNRF &
  DKPNet &
  0.1 &
  62.2 &
  104.7 &
  77.2 &
  137.5 &
  - &
  - &
  - &
  - &
  0.080 \\
SHA$\rightarrow$QNRF &
  DKPNet &
  0.5 &
  62.1 &
  103.9 &
  82.9 &
  149.7 &
  - &
  - &
  - &
  - &
  \textbf{0.078} \\
SHA$\rightarrow$QNRF &
  DKPNet &
  1.0 &
  62.5 &
  108.4 &
  81.2 &
  145.3 &
  - &
  - &
  - &
  - &
  0.085 \\  \hline
SHA$\rightarrow$QNRF$\rightarrow$SHB &
  DKPNet &
  0.1 &
  62.2 &
  104.7 &
  77.2 &
  137.5 &
  11.0 &
  19.8 &
  - &
  - &
  \textbf{0.040} \\
SHA$\rightarrow$QNRF$\rightarrow$SHB &
  DKPNet &
  0.5 &
  63.5 &
  109.6 &
  86.4 &
  147.5 &
  10.3 &
  17.3 &
  - &
  - &
  0.072 \\
SHA$\rightarrow$QNRF$\rightarrow$SHB &
  DKPNet &
  1.0 &
  62.5 &
  108.4 &
  81.2 &
  145.3 &
  10.7 &
  20.1 &
  - &
  - &
  0.043 \\\hline
SHA$\rightarrow$QNRF$\rightarrow$SHB$\rightarrow$NWPU &
  DKPNet &
  0.1 &
  65.5 &
  111.4 &
  92.5 &
  181.8 &
  8.7 &
  14.7 &
  84.4 &
  410.1 &
  0.042 \\
SHA$\rightarrow$QNRF$\rightarrow$SHB$\rightarrow$NWPU &
  DKPNet &
  0.5 &
  68.8 &
  113.9 &
  84.3 &
  160.1 &
  7.8 &
  12.2 &
  76.6 &
  364.2 &
  \textbf{-0.010} \\
SHA$\rightarrow$QNRF$\rightarrow$SHB$\rightarrow$NWPU &
  DKPNet &
  1.0 &
  67.0 &
  112.4 &
  84.8 &
  181.1 &
  11.0 &
  18.3 &
  80.0 &
  354.9 &
  0.079 \\\shline

\end{tabular}
}
\end{table*}

\subsection{Analysis of catastrophic forgetting}
As shown in Table~\ref{table: llcomparison}, we reproduce two of the classical lifelong learning methods and modify them to adapt to our crowd counting task, because most lifelong learning methods focus on the classification task, while crowd counting is a regression-like task. The average performances in past domains and unseen domains of our proposed FLCB method all surpass that of the LwF and EwC approaches. We compare the quantitative results between the baselines and our proposed method based on four benchmark models. The results in Table~\ref{table: catastrophic forgetting} clearly demonstrate that our method can remarkably alleviate the catastrophic forgetting phenomenon on all models with the lowest mMAE, mRMSE, and nBwT~(i.e., forgetting degree) under the domain-incremental training settings. We also report the model parameters and the Multiply-Accumulate Operations~(MACs) for each benchmark model. The forgetting degree in the intermediate process is also detailed in Table~\ref{results: length}. The results imply that the model will forget less and count better when more labeled datasets are involved in the lifelong learning process. This indicates that our framework can remember the old yet meaningful knowledge from the last well-trained model when handling the new domain dataset.

\begin{table*}[ht]
\caption{Generalization comparison of different training strategies on the unseen JHU-Crowd++ dataset. }
\label{table: generalization}
\centering
\begin{tabular}{l|cc|cc|cc|cc}
\shline
\multirow{2}{*}{\textbf{Model}} & \multicolumn{2}{c|}{\textbf{CSRNet}}     & \multicolumn{2}{c|}{\textbf{SFANet}}      & \multicolumn{2}{c|}{\textbf{DM-Count}}   & \multicolumn{2}{c}{\textbf{DKPNet}}     \\\cline{2-9}
                       & \textbf{MAE}           & \textbf{RMSE}           & \textbf{MAE}            & \textbf{RMSE}           & \textbf{MAE}           & \textbf{RMSE}           & \textbf{MAE}           & \textbf{RMSE}           \\\hline
JOINT                  & 103.2         & 320.0          & 115.5          & 347.6          & 96.3          & 320.3          & 89.8          & 318.7          \\
LwF*                   & 101.6         & 322.3          & 107.7          & 312.3          & 94.6          & \textbf{296.0} & 90.4          & 298.2          \\
FLCB                   & \textbf{92.9} & \textbf{305.1} & \textbf{102.2} & \textbf{311.3} & \textbf{82.5} & 298.5          & \textbf{84.8} & \textbf{264.8} \\
\shline
\end{tabular}
\end{table*}

\begin{table*}[ht]
\caption{Generalization comparison on the unseen JHU-Crowd++ dataset with self-distillation in different levels during the entire lifelong learning process. Here A, Q, B and N are the abbreviations for the name of four datasets~(SHA, QNRF, SHB, NWPU).}
\label{table: generalizationpromotion}
\centering
\resizebox{0.6\linewidth}{!}{
\begin{tabular}{cc|cc|cc|cc}
\shline
\multicolumn{2}{c|}{\textbf{Distillation}} &
  \multicolumn{2}{c|}{\textbf{A$\rightarrow$Q}} &
  \multicolumn{2}{c|}{\textbf{A$\rightarrow$Q$\rightarrow$B}} &
  \multicolumn{2}{c}{\textbf{A$\rightarrow$Q$\rightarrow$B$\rightarrow$N}} \\ \cline{1-8}
     \textbf{feature}&\textbf{output}& \textbf{MAE} & \textbf{RMSE} & \textbf{MAE} & \textbf{RMSE} & \textbf{MAE} & \textbf{RMSE} \\ \hline
\checkmark&  &102.6           &341.0            &93.2          &324.4           &87.1           &298.5          \\

&\checkmark  &106.7           &345.9            &102.3           &354.8            &90.4           &298.2            \\
\checkmark &\checkmark  &\textbf{96.2}       &\textbf{327.8}       &\textbf{90.5}       &\textbf{313.0}       &\textbf{84.8}       &\textbf{264.8}            \\ \shline

\end{tabular}}
\end{table*}

\begin{table*}[hb]
\caption{The experimental results of DKPNet with the synthetic-to-real training settings.}
\label{gccforgetting}
\centering
\resizebox{0.9\linewidth}{!}{
\begin{tabular}{l|cc|cc|cc|cc|c|c|c}
\shline
\multirow{2}{*}{\textbf{Method}} &
  \multicolumn{2}{c|}{\textbf{GCC-1}} &
  \multicolumn{2}{c|}{\textbf{GCC-2}} &
  \multicolumn{2}{c|}{\textbf{GCC-3}} &
  \multicolumn{2}{c|}{\textbf{GCC-4}} &
  \multirow{2}{*}{\textbf{mMAE}} &
  \multirow{2}{*}{\textbf{mRMSE}} &
  \multirow{2}{*}{\textbf{nBwT}}\\ \cline{2-9}
 &
  \textbf{MAE} &
  \textbf{RMSE} &
  \textbf{MAE} &
  \textbf{RMSE} &
  \textbf{MAE} &
  \textbf{RMSE} &
  \textbf{MAE} &
  \textbf{RMSE} &
   & \\ \hline
BASELINE &55.4  &131.3  &34.7  &82.8  &18.5  &53.3  &35.6  &74.9  &36.1  &85.6  &1.130\\
LwF*     &42.8  &104.5  &37.7  &108.5  &16.5  &43.1  &35.1  &70.6  &33.0  &81.7 & 0.378  \\
FLCB     &40.0  &95.4  &35.1  &100.5  &14.6  &34.5  &41.7  &82.5  &\textbf{32.8}  &\textbf{78.2} & \textbf{0.192} \\
\shline
\end{tabular}}
\end{table*}
\begin{table*}[ht]
    \caption{The test MAE and RMSE scores on the unseen ShanghaiTech PartB dataset after training synthetic GCC subsets.}
    \label{gccgeneralization}
        \centering
        \begin{tabular}{l | c c }
        \shline
        \textbf{Method} & \textbf{MAE} & \textbf{RMSE} \\ \hline
        JOINT & 22.8 & 30.6 \\
        CycleGAN~\cite{zhu2017unpaired}  & 25.4 & 39.7\\
        SE CycleGAN~\cite{wang2019learning} & 19.9 & 28.3\\
        FLCB & \textbf{16.1} & \textbf{25.0}\\\shline
       \end{tabular}
\end{table*}

\begin{figure*}[ht]
\centering
   \includegraphics[width=0.9\linewidth]{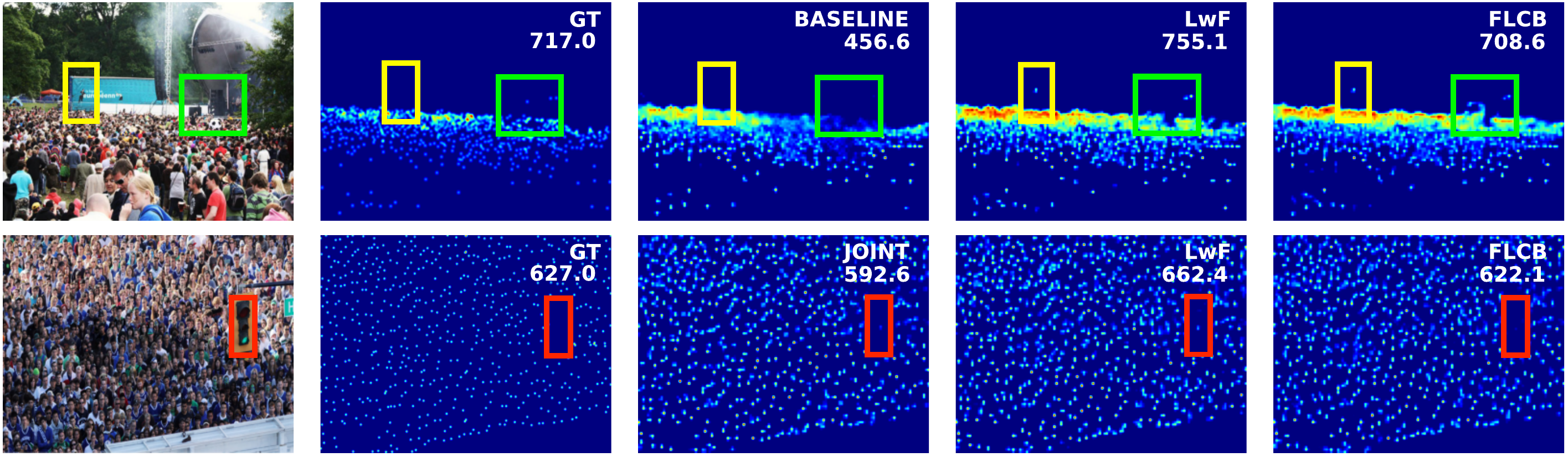}
   \caption{The visualization results of different training paradigms. The top row shows the predictions and compares the forgetting degree on the first training dataset (SHA), while the bottom row illustrates the predictions and compares the generalization ability on the unseen dataset (JHU). Red: FLCB can correctly discriminate the non-human objects like traffic lights. Green: FLCB may be affected by the background noise such as loudspeakers. Yellow: FLCB may not deal with the missing annotations well which is not the key research point in our work.}
   \label{figures: visualization}
\end{figure*}

\subsection{Effect of hyper-parameter\texorpdfstring{$\lambda$}{}}
The proposed balanced domain forgetting loss~(BDFLoss) is composed of optimal transport counting loss and self-distillation loss. The hyper-parameter $\lambda$ plays a dominant role in our proposed BDFLoss to control how much previously learned meaningful knowledge should be retrained when learning on new domain data. In other words, the hyper-parameter $\lambda$ is a trade-off between model plasticity and stability. The greater the value of $\lambda$ is, the more attention should be paid to leveraging the distilled knowledge. If $\lambda$ is equal to 0, it degenerates to the vanilla sequential training without any constraints of previous knowledge. We just empirically choose the $\lambda=0.5$ to conduct our main experiments in this paper. In this subsection, we also investigate whether different $\lambda$ values will have a visible effect on forgetting. The extensive results demonstrate that the $\lambda=0.5$ is a reasonable choice as a trade-off of model plasticity and stability in Table~\ref{results: lambda}.

\subsection{Analysis of model generalization}
\subsubsection{Real-to-real generalization}To build a robust model for better crowd counting, we also expect that the model can obtain acceptable performance among unseen domains, because labeling crowd images is extremely expensive and time-consuming in the real world. After the ultimate models converge, we test them directly on the unseen JHU-Crowd++ dataset in Table~\ref{table: generalization}. 
Note that the images from JHU-Crowd++ are never trained during the process of lifelong learning. Our proposed FLCB can achieve lower prediction errors in terms of MAE and RMSE over the unseen dataset, indicating a stronger generalization ability compared with the joint training strategy. Furthermore, taking the DKPNet as an example, we delve into the ablation study of different layers for distillation in the intermediate lifelong learning process. Every time a new incoming dataset is finished training, the model will be evaluated on the unseen dataset. The results, shown in Table~\ref{table: generalizationpromotion}, illustrate that its performance is boosted progressively with incremental data from different domains. It also indicates that the model can count better on the unseen domain under the mutually complementary interaction of both feature-level and output-level distillation. Training in different orders may achieve fluctuating performance in unseen domains. We also present the results in Appendix~\ref{appendix: training orders} because it could be related to curriculum learning, which is not our main focus in this paper.
\subsubsection{Synthetic-to-real generalization}
Considering the data privacy and some ethical policies~(\textit{i.e.}, the real-world training images may be unobtainable), we conduct the same lifelong training settings training on the synthetic crowd dataset (GCC)~\cite{wang2019learning} and investigate the generalization on the unseen real-world dataset~(ShanghaiTech PartB). The GCC dataset is collected from the GTA5 game environment, which contains 15,212 synthetic images with diverse scenes. The synthetic dataset can provide precise but not time-consuming annotations for training. We split the GCC synthetic dataset into four subsets to mock the same lifelong training settings. The forgetting phenomenon among incremental synthetic subsets is still analyzed in Table~\ref{gccforgetting}, as well as the generalization performance on the unseen dataset. After obtaining the ultimate model, our FLCB benchmark achieves the lowest mMAE, mRMSE, and nBwT among previously seen datasets and decent performance on the unseen real-world dataset. Furthermore, the generalization experimental results in Table~\ref{gccgeneralization} verify the superiority of our proposed benchmark. 

In summary, our proposed lifelong crowd counting benchmark~(FLCB) can help the crowd counters forget less and count better to sustainably handle the multiple-domain crowd counting using a single model, which indicates it has a promising potential to tackle the more complicated scenes in the future.

\subsection{Visualization results}
To make a more qualitative comparison, we visualize the prediction density maps under different training strategies. As illustrated in Figure~\ref{figures: visualization}, we can discover that the sequential training methods will achieve terrible performance among old domains after training images from a new domain. Our proposed lifelong crowd counting benchmark can estimate on both seen and unseen datasets more accurately and outperforms other training paradigms.

\subsection{Discussions}
\noindent\textbf{Limitations:} In this paper, we attempt to develop a single model to handle the incremental datasets from different domains for better lifelong crowd counting. Judging from both quantitative and qualitative results, our proposed FLCB does well in achieving a trade-off performance from all domain datasets compared with other methods. However, there are still some limitations to be discussed that may drive the future research directions in lifelong crowd counting. On one hand, according to the visualization results, our proposed FLCB method seems to have difficulty in dealing with the missing annotations~(yellow bounding boxes) and background noises~(green bounding boxes), like the loudspeaker box in Figure~\ref{figures: visualization}. On the other hand, we do not integrate any replay-based strategies into our experiments considering the training time and storage overhead. Efficient data sampling strategies and replay-based approaches may boost the lifelong crowd counting which deserves to be investigated in the future.

\noindent\textbf{Lifelong learning v.s. Self-supervised Learning:} We would like to discuss the lifelong learning and self-supervised learning from a pretraining perspective. They share something in common that is expected to lay the foundation for the Artificial General Intelligence~(AGI). Recent literature~\cite{caron2020unsupervised,he2020momentum,huang2022learning,chen2020simple,grill2020bootstrap,niu2022self,niu2020suppression,niu2021spice,niu2022unsupervised} shows the power of self-supervised learning as a novel pretraining paradigm to empower multiple downstream tasks. To an extent, lifelong learning could also be regarded as a kind of pretraining method, because it learns the shared knowledge and general representations to boost performance. However, lifelong learning usually requires labeled data for training to enhance the model capacity, whereas self-supervised learning does not. From our perspectives, both types fo learning could provide a good pretrained network or initialization for other domain datasets' or downstream tasks' training, and lifelong learning may empower self-supervised learning in the future.

\section{Conclusion}
Our work proposes a domain-incremental self-distillation learning benchmark for lifelong crowd counting to try to resolve the catastrophic forgetting and model generalization issues using a single model when training new datasets from different domains one after another. With the help of the BDFLoss function that we designed, the model can effectively forget less and count better during the entire lifelong crowd counting process. Additionally, our proposed metric nBwT can be used to measure forgetting degree in future lifelong crowd counting models. Extensive experiments demonstrate that our proposed benchmark has a lower forgetting degree over the sequential training baseline and a stronger generalization ability compared with the joint training strategy. Our proposed method is a simple yet effective way to sustainably handle the crowd counting problem among multiple domains using a single model with limited storage overhead when the newly available domain data arrive. It can be incorporated into any existing backbone as a plug-and-play training strategy for better crowd counting in the real world. Although our work considers crowd counting, further, the proposed framework has the potential to be applied in other regression-related image or video tasks.

\bibliographystyle{unsrt}  
\bibliography{references}

\clearpage
\appendix
\section*{Appendix}
\section{Domain Concept and Gaps of Different Datasets}\label{appendix: domain concept}
\begin{figure*}[ht]
  \includegraphics[width=\textwidth]{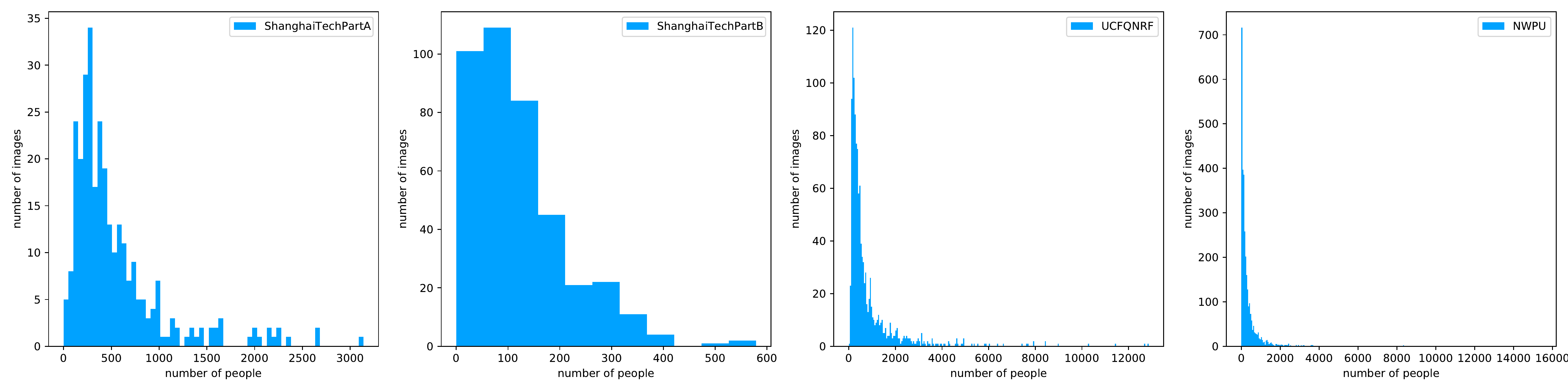}
\caption{The data distributions of four benchmark datasets.}
  \label{fig:datadistribution}
\end{figure*}

A domain $D$ consists of two components: a feature space $X$ and a marginal probability distribution $P(x)$, (\ie, $D={X, P(x)}$), according to the definition in~\cite{weiss2016survey}. It implies that if two domains ($D_A$ and $D_B$) are different they may either have different feature spaces ($X_A \neq X_B$) or different marginal probability distributions ($P(X_A) \neq P(X_B)$). In crowd counting tasks, on one hand, off-the-shelf datasets are captured from different cameras or different scenarios like streets, museums, or gyms, so the data distributions are different. On the other hand, from the Bayesian perspective (\ie, $P(x) = P(c) / P(c|x)$, where $c$ is the number of people and $x$ is the given crowd image), the marginal probability distribution $P(x)$ of each dataset is also different.  The $P(c|x)$ is our single learning model $f(\cdot)$ which is fixed capacity and maps from the input images $x$ to the estimated count number $c$. The $P(c)$ represents the population density of each dataset, and varies from dataset to dataset, as shown in the Figure~\ref{fig:datadistribution}. Thus, there exist the domain gaps among these crowd counting datasets with different $P(x)$. This is our further theoretical analysis on the core domain concept.

Specifically, the ShanghaiTech PartA dataset is collected from the Internet with the highly-variant density distribution ranges from 33 to 3,139 pedestrians per image. As a larger crowd counting dataset, the UCF-QNRF dataset includes 1,535 images collected from several image search engines like Google Image Search, Flicker, and so on. In contrast to other available public datasets, the NWPU-Crowd dataset is a generally more extensive and more crowded dataset annotating heads from 0 to 20,033 per image, which first introduces negative samples like extremely high-density images and images containing zero people. As shown in Figure~\ref{fig:datadistribution}, the ShanghaiTech PartB dataset contains fewer people per image (123 people on average) compared to the other three datasets (501,815,418 people on average) used for experiments in our paper. Such a prominent domain shift problem motivates us to investigate the catastrophic forgetting and generalization issues in the lifelong crowd counting task in this study.

In the proposed lifelong crowd counting task, data comes from non-stationary and changing distributions, which means $P_{t_i}(X,Y) \neq P_{t_j}(X,Y)$, where $t_{i}$ and $t_{j}$ represent different time-steps $t$, and $X$ and $Y$ are the crowd images and their corresponding ground truth density maps (labels). Different from typical crowd counting, the distribution shifting problem raises a challenge in lifelong crowd counting. At different time-steps, the marginal distribution of crowd images $X$ shifts among different datasets, while the generation of the ground truth remains the same, which is $P_{t_i}(X) \neq P_{t_j}(X)$ and $P_{t_i}(Y|X) = P_{t_j}(Y|X)$. To tackle the lifelong crowd counting task, we specifically present one unseen dataset~(JHU-Crowd++) that focuses on the generalization of crowd counting, which is much larger than any of the existing datasets, including seen and unseen domains.

To construct the seen domains, we organize four popular crowd counting datasets, including ShanghaiTech PartA, ShanghaiTech PartB, UCF-QNRF, and the NWPU-Crowd. The training set (7,092 images) comes from the four different datasets' training sets. The current domain performance and forgetting degree are evaluated on the corresponding test sets from the four datasets. To illustrate the model generalization ability, we choose only the test set of the JHU-Crowd++ dataset (1,600 images) as the unseen domain dataset, because it has a more significant counts span. For a fair comparison with other training paradigms, none of the images in JHU-Crowd++ dataset are trained during the lifelong learning process.

\begin{table*}[ht]
\caption{The forgetting degree comparison results  with different training orders.}
\label{results: orderforgetting}
\resizebox{\linewidth}{!}{
\begin{tabular}{l|cc|cc|cc|cc|c|c|c}\shline
\multirow{2}{*}{\textbf{Training order}} &
  \multicolumn{2}{c|}{\textbf{SHA}} &
  \multicolumn{2}{c|}{\textbf{QNRF}} &
  \multicolumn{2}{c|}{\textbf{SHB}} &
  \multicolumn{2}{c|}{\textbf{NWPU}} &
  \multirow{2}{*}{\textbf{mMAE}} &
  \multirow{2}{*}{\textbf{mRMSE}} &
  \multirow{2}{*}{\textbf{nBwT}} \\\cline{2-9}
 &
  \textbf{MAE} &
  \textbf{RMSE} &
  \textbf{MAE} &
  \textbf{RMSE} &
  \textbf{MAE} &
  \textbf{RMSE} &
  \textbf{MAE} &
  \textbf{RMSE} &
   &
   &
   \\\hline
SHA$\rightarrow$QNRF$\rightarrow$SHB$\rightarrow$NWPU(BASELINE) &
  92.9 &
  157.8 &
  100.1 &
  179.4 &
  7.7 &
  12.4 &
  90.0 &
  393.6 &
  72.7 &
  185.8 &
  0.371 \\
SHA$\rightarrow$QNRF$\rightarrow$SHB$\rightarrow$NWPU(FLCB) &
  68.8 &
  113.9 &
  84.3 &
  160.1 &
  7.8 &
  12.2 &
  76.6 &
  364.2 &
  \textbf{59.4} &
  \textbf{162.6} &
  \textbf{-0.010} \\\hline
NWPU$\rightarrow$QNRF$\rightarrow$SHA$\rightarrow$SHB(BASELINE) &
  124.9 &
  229.0 &
  240.1 &
  435.8 &
  7.4 &
  12.5 &
  218.2 &
  826.5 &
  147.7 &
  376.0 &
  1.576  \\
NWPU$\rightarrow$QNRF$\rightarrow$SHA$\rightarrow$SHB(FLCB) &
  62.3 &
  108.0 &
  78.8 &
  138.6 &
  10.7 &
  20.2 &
  94.8 &
  417.5 &
  \textbf{61.7} &
  \textbf{171.1} &
  \textbf{0.043}  \\\hline
QNRF$\rightarrow$SHA$\rightarrow$SHB$\rightarrow$NWPU(BASELINE) &
  87.1 &
  162.6 &
  107.1 &
  212.8 &
  10.1 &
  16.1 &
  100.1 &
  462.1 &
  76.1 &
  213.4 &
  0.432 \\
QNRF$\rightarrow$SHA$\rightarrow$SHB$\rightarrow$NWPU(FLCB) &
  61.3 &
  104.8 &
  84.2 &
  149.7 &
  10.3 &
  17.9 &
  83.9 &
  377.8 &
  \textbf{59.9} &
  \textbf{162.6} &
  \textbf{-0.001} \\

\shline
\end{tabular}
}
\end{table*}

\begin{table}[ht]
\caption{The generalization comparison results with different training orders on the unseen JHU-Crowd dataset.}
\label{results: ordergeneralization}
\centering
\begin{tabular}{l|cc}\shline
\multirow{2}{*}{\textbf{Training mode}} &
  \multicolumn{2}{c}{\textbf{JHU}} \\\
 &
  \textbf{MAE} &
  \textbf{RMSE} \\\hline
  ShanghaiTech PartA &
  106.0 &
  338.3 \\
ShanghaiTech PartB &
  154.4 &
  530.2 \\

UCF-QNRF &
  97.8 &
  315.9 \\
NWPU-Crowd &
  94.5 &
  323.4 \\\hline
JOINT &
  89.8 &
  318.7 \\\hline
NWPU$\rightarrow$QNRF$\rightarrow$SHA$\rightarrow$SHB &
  \textbf{87.2} &
  \textbf{287.5} \\
SHA$\rightarrow$QNRF$\rightarrow$SHB$\rightarrow$NWPU &
  \textbf{84.8} &
  \textbf{264.8} \\
QNRF$\rightarrow$SHA$\rightarrow$SHB$\rightarrow$NWPU &
  \textbf{83.4} &
  \textbf{264.8} \\
\shline
\end{tabular}
\end{table}

\section{Effect of Different Training Orders}\label{appendix: training orders}
\noindent\textbf{Forgetting Degree Analysis.}\quad
We compare the results of our proposed framework with different training orders and the corresponding baseline models in Table~\ref{results: orderforgetting}. The results clearly show that our method can mitigate the forgetting phenomenon in the lifelong crowd counting process compared with the vanilla sequential training strategy under the circumstances of different training orders.

\noindent\textbf{Generalization Analysis.}\quad
To avoid the generalization performance improvement caused by a particular training order, we also conduct the same experiments with different training orders of four benchmark datasets. As shown in Table~\ref{results: ordergeneralization}, the model still achieves outstanding performance on the unseen JHU-Crowd++ dataset in contrast with the single-domain training settings and joint training strategy. Compared with joint training and individual training, the results clearly verify the effectiveness of our proposed domain-incremental self-distillation learning framework for consistently strengthening the model generalization ability consistently with different training orders.

\end{document}